\def\year{2020}\relax
\documentclass[letterpaper]{article} 
\usepackage{aaai20}  
\usepackage{times}  
\usepackage{helvet} 
\usepackage{courier}  
\usepackage[hyphens]{url}  
\usepackage{graphicx} 
\usepackage{graphicx}  
\usepackage{times}
\usepackage{latexsym}
\usepackage{algorithm}
\usepackage[noend]{algpseudocode}
\usepackage{amsmath,amssymb}
\usepackage{bm}
\newcommand{\citet}[1]{\citeauthor{#1} \shortcite{#1}}
\newcommand{\citep}{\cite}
\newcommand{\citealp}[1]{\citeauthor{#1}, \citeyear{#1}}

\title{Sample-Efficient Model-based Actor-Critic for an Interactive Dialogue Task}
\author {Katya Kudashkina\textsuperscript{1,}\textsuperscript{2},
Valliappa Chockalingam\textsuperscript{3},
Graham W. Taylor\textsuperscript{1,}\textsuperscript{2},
Michael Bowling\textsuperscript{3,4}\\\\
\textsuperscript{1}{University of Guelph}, 
\textsuperscript{2}{Vector Institute for Artificial Intelligence},\\
\textsuperscript{3}{University of Alberta},
\textsuperscript{4}{Alberta Machine Intelligence Institute}\\\\
A Preprint
        }
 
\begin{document}

\maketitle

\begin{abstract}
Human-computer interactive systems that rely on machine learning
are becoming paramount to the lives of millions of people who use digital assistants on a daily basis.
Yet, further advances are limited by the availability of data and the cost of acquiring new samples.
One way to address this problem is by improving the sample efficiency of current approaches.
As a solution path, we present a model-based reinforcement learning algorithm
for an interactive dialogue task.
We build on commonly used actor-critic methods,
adding an environment model and planner that augments a learning agent to learn the model of the environment dynamics.
Our results show that, on a
simulation that mimics the interactive task,
our algorithm requires 70 times fewer samples, compared to the baseline of commonly used model-free algorithm, and
demonstrates 2~times better performance asymptotically.
Moreover, we introduce a novel contribution of computing a soft planner policy and
further updating a model-free policy yielding a less computationally expensive model-free agent as good as the model-based one.
This model-based architecture serves as a foundation that can be extended to other
human-computer interactive tasks allowing further advances in this direction.
\end{abstract}

\section{Introduction}
Intelligent assistants that amplify and augment human cognitive and physical abilities have a paramount influence on society.
The ubiquitous nature of intelligent assistants raises their economic significance. Voice personal assistants, in particular, are on the rise. These are systems such as Amazon Alexa, Google Personal Assistant, Microsoft Cortana, Apple Siri, and many others, that assist people with a myriad of tasks: gathering information, inquiring about the weather,
making purchases, booking appointments, or dictating and editing documents.
Such technology that helps in performing tasks through
a dialogue between a user and a system
is often referred to as~\textit{voice-assistive}.

We center our attention on a particular class of voice-assistive systems: voice document-editing.
Work on document editing via voice goes
back more than three decades~\cite{ades1986voice}.
Despite recent progress, voice document-editing and voice-dictation systems
are still in a primitive form.
For example, in some cloud-based document editors, we can type,
format, and edit using {only specific commands}:
``Select line'' in order to select a line, or ``Go to the next character'' to advance the cursor.
The lack of sophistication in today's voice-editing systems is due to the
difficulty of training them.
They require either a large, diverse training dataset of general document-editing
dialogues or hours of online human-machine interactions.
Such datasets do not currently exist and online interactions are impractical---
a substantial amount of training is needed before the agent can overcome
users' frustration with a system that has not been sufficiently trained.

General methods, such as supervised learning, that scale with increased computation, continue to build knowledge into our agents.
However, this built-in knowledge is not enough when it comes to conversational AI agents assisting users, especially in complicated domains.
Supervised learning is an important kind of learning, but alone it is not adequate for learning from interaction.
In contrast, reinforcement learning provides an opportunity for an intuitive human-computer interaction.

The key contribution of this paper is a sample-efficient
model-based reinforcement learning
algorithm that can make best use of limited datasets and online learning.
Its novelty is in computing a soft planner policy and updating a model-free policy. By integrating such an update with the rest of the algorithm, we can deploy a model-free agent that performs better than an agent trained without model-based components.
We show that our algorithm is 70 times more sample-efficient than the one trained with a commonly
used model-free actor-critic method that is already known for
its good performance across many domains.
In addition, asymptotically our algorithm demonstrates twice the performance in comparison to
the model-free actor-critic baseline.

The intent of our algorithm is to address sample efficiency
in {\it real-world human-computer interactive systems}, where
the reward function truly depends on user feedback.
Thus, we validate the algorithm
on a selected challenging task
that carries this property and encompasses user interactions with an imperfect transcription system--a voice
document-editing task.
The work demonstrates the benefits of a model and planner with a novel soft-planner policy
and shows how to fine tune a model-free policy and value function.
To establish our algorithm we
compare it to the current state-of-the-art implementations
for such systems: model-free actor-critic methods.

\section{Related Work \& Background}
Reinforcement Learning (RL) is a framework that allows an agent to learn by
interacting with an environment.
The agent maps situations to actions with the goal of
maximizing some numerical reward signal.
It is not surprising that RL techniques have found wide use in a number
of dialogue system implementations~\cite{walker1978understanding,singh2000reinforcement,shah2016interactive,li2016deep,dhingra2016towards}.
Below we describe the key RL concepts and terminology closely following the~\citealp{sutton2018reinforcement} conventions for notation and definitions.

Commonly, RL problems are cast as sequential decision-making problems modelled
as~\textit{Markov Decision Processes} (MDPs).
In an MDP, an agent and an environment interact in a sequence of discrete steps.
The state at time-step $t$ is denoted $S_t \in \mathcal{S}$, where $\mathcal{S}$
is the set of all possible states.
In cases of partial observability, the environment emits
only~\textit{observations} $O_t$---a signal that depends on its state, but carries only partial information about it.
The observations may not contain all information necessary for
predicting the future or optimal control (i.e., learning policies to attain a large reward.)
In this case we operate within the framework known as~\textit{Partially Observable MDPs} (POMDPs).

An action that the agent takes at the current state is defined as
$A_t \in \mathcal{A}(s)$, where $\mathcal{A}(s)$ is the set of all actions possible
from state $s$.
Action selection is driven by the information that state $s$ provides.

The policy $\pi$ is a probability distribution over all possible
actions $a \in \mathcal{A}(s)$ at a given state $s$:
$p(a|s) = \text{Pr}\{A_t=a| S_t =s\}$.
Once an action is taken, the agent receives a numerical \textit{reward},
$R_{t+1} \in \mathbb{R}$ and transitions to the next state $S_{t+1} \in \mathcal{S}$,
according to the transition probabilities of the~\textit{environment's} dynamics:
$ p(r, s' | s, a) = \text{Pr} \{ R_ {t+1} = r, S_{t+1} = s' | S_t=s, A_t =a\}$,
producing a sequence:
\begin{equation}
\label{sequence}
    S_0, A_0, R_1,S_1, ..., S_{T-1}, A_{T-1}, R_T, S_T
\end{equation}
where $T$ is the final timestep for an episodic task or infinity for a continuing task.
The goal of the agent is to learn the~\textit{optimal policy} $\pi^*$
which maximizes the total~\textit{expected discounted return}:
\begin{equation}
    \resizebox{0.425\textwidth}{!}{%
    $
    G_t = R_{t+1} + \gamma R_{t+2} + \gamma^2 R_{t+3}+ ...
    = \sum_{k = 0}^{T-t-1} \gamma^k R_{t + k +1} $
    }
\end{equation}
where $\gamma$ is a parameter, $ 0 \leq \gamma \leq 1$ called
the~\textit{discount rate}.

In~\textit{value-based methods}, we compute~\textit{value functions}.
In prediction-type problems, it is common to estimate the~\textit{state-value function}
as $v_{\pi} = \mathbb{E}_{\pi}[G_t|S_t = s]$ when
starting in state $s$ and following policy $\pi$.
Prediction problems are also referred to as~\textit{policy evaluation}.
In control-type problems, we estimate the~\textit{action-value function} from
which we derive the policy:
$q_{\pi}(s,a) = \mathbb{E}_{\pi}[G_t|S_t = s, A_t = a]$.
In~\textit{policy-based methods}, we directly optimize and compute the policy.

The first suggestions of formalizing
MDPs for dialogue goes back forty years,
followed by work using RL~\cite{walker1978understanding,biermann1996composition,levin1997learning}.
An example of an early end-to-end dialogue system that used RL
is the RLDS software tool developed by~\citealp{singh2000reinforcement} at AT\&T Labs.
NJFun was another pioneering system developed by~\citealp{litman2000njfun}
that used RL.
NJFun was created as an MDP with a manually designed state space
for the dialogue.

Dialogue systems have rapidly advanced their performance
in the last few years,
following the introduction of the sequence-to-sequence paradigm by
\citealp{sutskever2014sequence} which transformed machine translation among
other natural language processing tasks.
Much of this progress is now attributed to the
increasing availability of large amounts of data and computational power via deep learning
techniques combined with RL.
Most current dialogue systems that apply RL combine
value-based and policy-based methods
within the \textit{actor-critic} framework described next.

\subsection{Actor-Critic Methods}
\label{AC-background}
Actor-critic algorithms~\cite{barto1983neuronlike} are methods that combine
value-based and policy-based methods.
The `actor' refers to the part of the agent responsible for producing the policy, while the `critic' refers
to the part of agent responsible for producing a value function.
Actor-critic is a subset of policy gradient algorithms, which we describe next.

In policy gradient methods, we learn a~\textit{parameterized policy}:
\begin{equation}
    \pi_{\theta}(a|s,\bm{\theta}) = \text{Pr}\{A_t = a| S_t =s, \bm{\theta}_t = \bm{\theta}\},
\end{equation}
where $\bm{\theta} \in \mathbb{R}^{d'}$, and $d'$ is the dimensionality of $\bm \theta $.
The goal of the learning agent is to maximize some scalar performance measure $J(\bm{\theta})$
with respect to the policy parameters.
Policy gradient methods find a locally
optimal solution to the problem of maximizing the objective  $J$ and work by applying gradient ascent:
\begin{equation}
    \bm{\theta}_{t+1} = \bm{\theta}_t + \alpha \widehat{\nabla_\theta J (\bm{\theta}_t)},
\end{equation}
where $\widehat{\nabla J (\bm{\theta}_t)} \in \mathbb{R}^{d'}$ is a
stochastic estimate whose expectation approximates the gradient of the performance
measure with respect to its argument $\bm{\theta}_t$.
The classic variant of policy gradient methods is REINFORCE~\cite{williams1992simple}, where
the gradient of the parameters of the objective function is
\begin{equation}
    \nabla_{ \theta}J(\bm \theta) =
    \mathbb{E}[\nabla_{\theta} \log (\pi_{\theta}(a|s,\bm{\theta}))G_t].
\end{equation}
In REINFORCE, the agent simply collects transitions from an episode using the current policy
$\pi$ and then uses it to update the policy parameters.
The gradient estimation technique can have high variance when using only an actor. Hence, a large number of samples may be needed for the policy to converge.
To solve this problem, actor-critic methods aim to reduce this variance
by using an actor and a critic.
The critic's approximate value of a state  $\hat v(s, \bm w)$ is parameterized by $\bm w$,
where $\bm w \in \mathbb{R}^{d}$ and $d$ is the number of components of the parameter vector.
For the critic update, in the simplest case of 1-step Temporal Difference learning, we perform the following:
\begin{equation}
\label{eq-critic-update}
\begin{split}
    \delta \gets R_{t+1} + \gamma\hat v (S_{t+1}, \bm w_{t}) - \hat v(S_{t}, \bm w_{t})\\
    \bm w_{t+1} \gets \bm w_{t} + \alpha_{w} \delta \nabla_{w}\hat v(S_{t}, \bm w_{t}).
\end{split}
\end{equation}

In this work, we focus on the~\textit{Advantage Actor-Critic} (A2C) method~\cite{mnih2016asynchronous},
which make updates to policies using the~\textit{advantage function}:
$\mathbb{A}(s, a, \bm w^q, \bm w^v) = \hat{q}(s, a, \bm w^q) - \hat{v}(s, \bm w^v)$,
where $\hat{q}(s, a, \bm w^q)$ is an estimate of the value of the state-action pair $(s, a)$ and $\mathbb{A}$ is
parameterized by $\bm w^q$ and $\bm w^v$.
To update $\bm w^v$ and $\bm w^q$ (using action-value estimates instead of state-value estimates) we can use temporal difference updates as described in Equation~\ref{eq-critic-update}.
To update $\bm \theta$, instead of using $G_t$, as in REINFORCE, we use a policy gradient update with $\mathbb{A}$:
\begin{equation}
\begin{split}
    \mathbb A(S_t, A_t, \bm w^q_t, \bm w^v_t) = \hat{q}(S_t, A_t, \bm w^q_{t}) - \hat{v}(S_{t+1}, \bm w^v_{t})\\
    \bm \theta_{t+1}  = \bm \theta_{t} +
                \alpha_{\theta} \mathbb A(S_t, A_t, \bm w^q_t, \bm w^v_t) \nabla \text{ln} \pi_{\theta} (A_t | S_t, \bm \theta_t) .\
\end{split}
\end{equation}
In practice, instead of using a separate action-value network $\hat{q}(s, a, \bm w^q)$, empirically observed truncated $n$-step returns are used to calculate an estimate of the advantage:
$\hat{\mathbb A}(S_t, A_t, \bm w^v_t) = R_{t + 1} + \gamma R_{t + 2} + ... + \gamma^{n - 1} R_{t + n} +
\gamma^n \hat v(S_{t + n}, \bm w^v_t) - \hat v(S_t, \bm w^v_t)$, where  episode trajectory rollouts are performed for $n$ steps during which the value function remains constant.

\subsection{Model-based Reinforcement Learning}
\label{sec-model-based-background}
Learning the dynamics of the environment to augment the state or for planning---\textit{model-based RL}---is a
promising approach for improving sample efficiency~\cite{kaiser2019model}.
In this particular paradigm of model-based reinforcement learning, knowledge is represented in two parts: a state part and a dynamics part~\cite{sutton2019MBRL}.
The state part can be formulated as a~\textit{state-update function} which aims to capture all the information needed for predicting and controlling future state and rewards.
In a classic MDP with full observability, a state, referred to as Markov state, contains all necessary information to predict future states and to take action.

The second part of model-based reinforcement learning, is learning a model for the dynamics of the environment,
captured by an~{\it environment model}.
A common approach for this has been to learn an environment model online
by predicting future states and rewards and comparing them to the ones the agent sees later.
An environment model can be learned by the agent during human-computer interactions
to capture the dynamics of interactions.
These dynamics are everything that the agent needs to know about the environment.
Environment models help the agent in making informed action choices
by allowing it to compute~\textit{possible} future outcomes when considering different actions.
These actions may include actions that the agent does not take and this process is called~\textit{planning}.
Planning is particularly important because it allows the agent to
learn not only from the actions the agent takes,
but also from the actions that the agent does not select.
Planning is exactly what allows a model to sweep through all actions, and propagate information back making learning easier and faster---in other words, sample-efficient.

However, current methods in dialogue systems that are based on policy-gradient methods
do not include critical components of model-based RL
that allow an agent to learn the dynamics of the environment it operates in,
in our case, dynamics of a dialogue environment.
A number of works combine policy-gradient
methods and supervised learning~\cite{he2015deep,li2016deep,kaplan2017beating,stupidRLpaperondialogues}, or
take a different approach to improve sample complexity using imitation learning
\cite{lipton2018bbq}.
The closest to using a model-based RL approach is the work of~\citealp{zhao2016towards}.
Their end-to-end architecture is close to the Dyna architecture~\cite{sutton1991dyna}.
However, their additional model employs a known transition function from a database
and not a model that is learned through interactions.
One may compare their methods to MBPO method~\cite{janner2019trust}
that also uses offline data while our method focuses on {\it online learning}.

Our algorithm allows the agent to learn a model online from interactions and 
removes the complexities of needing additional elements,
such as ``successful trajectories''~\cite{lipton2018bbq}
or data augmentation~\cite{goyal2019using}.

To demonstrate the advantages of our model-based RL algorithm
within dialogue tasks, we selected a particular task of
voice document-editing driven by two primary reasons.
The first reason is that the voice document-editing application
can be simulated while
\textit {accurately reflecting
real-life scenarios} of voice document-editing task
which permits the development and evaluation of models without user interaction.
The second reason is that, in general, we think that model-based RL
is a useful paradigm for human-computer interactive tasks with limited data.
Voice document-editing, an example of such tasks, is a first step in this direction.
Additionally, this task is particularly difficult for supervised approaches
given the variability of users' behavior.
The next section describes the task in detail.

\section{Voice Document-Editing Task}
Voice document-editing systems could be incredibly powerful
if they could fully process a dictation in free-form language.
Anyone who creates documents could have a voice-assistant helping
them with document editing in a timely and efficient manner.
People could create and edit documents on-the-go: delete and insert words;
create or edit itemized lists; 
change the order of words, paragraphs, or sentences; convert one tense to another;
add signatures or salutations;
format text; fix
grammar; or fine-tune the style.
Often, a user simply wants to delete or correct what they dictated a moment ago.
The reason for this is that current speech recognition systems are imperfect.
For example, a human might mean to say:  ``Good morning, George.
I hope you have been well'', yet the dialogue system transcribes:
``Good morning, George. I hope you pin well.''
In this case, the user wishes to delete and re-dictate the final part of the sentence.
In this paper, we focus on a specific problem in voice document-editing systems that manifests from the state of current speech recognition systems.
Specifically, today's systems do not allow the user to correct errors via voice.

We proceed with a setting where misrecognized words are
at the end of the sentence, which is common in real-life scenarios.
As soon as a dictation tool displays incorrectly processed words,
a user sees it right away and stops dictating.
In our setting, the user says ``No'' if she is not satisfied with the
words at the end of the sentence that are displayed on the screen during dictation.
When the user says ``No'', then ``No'' is treated as a keyword for
the agent to take an action: to identify which words were not recognized properly
and to delete them.
If the agent deletes too few words, the user can say ``No'' again
and the process is repeated until the user is satisfied.
If the agent deletes too many words, the
user has to repeat the accidentally-deleted words before continuing dictation.
The agent's goal is to identify the correct number of words to delete in the least number of steps.
We formalize our setting using RL~concepts.\\
\textit{Environment}: An environment is comprised of a text
that is transcribed from a user's dictation
and the user speaking to the agent.\\
\textit{Action}: An action is the number of words to delete.\\
\textit{Interaction}: An interaction is an interactive process between the user
and the agent. This interaction starts the moment
a user says ``No'' for the first time, after dictating a new sequence of words,
and terminates in one of the following two cases: 1) If the user is satisfied with
the sentence and last edits (last action from the agent); or
2) If the agent deletes too many words.
In the latter case, the user has to repeat some words that were previously dictated, because
the agent deleted some well-recognized words by mistake.
The satisfaction of the user is indicated by the fact that the user simply continues
dictation and does not say ``No'' after the agent takes an action.\\
\textit{Intent}: An intent is an integer representing the number of words
that the user wishes to correct.\\
\textit{Word sequence $w_t$}: A word sequence is a sequence of words that is shown on
the screen at time step $t$.\\
\textit{Speech transcription $\psi_t$}: A speech transcription is a new sequence of words dictated by a human
at time step $t$. It is an output from a speech-to-text system.\\
\textit{Observation $O_t$}: An observation is a combination of
a word sequence $w_t$ and speech $\psi_t$ that the agent observes at time step $t$.
An observation is emitted from the environment and serves as an input to a state-update function.\\
\textit{State $S_t$}: A state is a representation of the entire interaction's history
at time $t$.
The state is computed by the state-update function and
it accumulates all the information since the beginning of an interaction: information about
the observation, and the previous action and state.\\
\textit{Reward}: The reward for each action is $0$ if the action corresponds to the intent.
If the action is greater than the intent, the agent is penalized by the number
of words $b_t$ that were not noisy but deleted by the agent incorrectly.
The reward then is $-1 \cdot b_t$.
If the agent undershoots by deleting fewer words than the intent,
the reward is $-1 \cdot m_t$ where $m_t$ is
the number of steps within an interaction
and is equal to the number of time the user already said ``No''.
This reward function reflects user (dis)satisfaction
penalizing for longer time to task completion and overshoots
that will require further interactions.\\
Consider the previous example where the user says ``Good morning, George.
I hope you have been well'', yet the dialogue system transcribes
``Good morning, George. I hope you pin well.''
The initial intent here is 2, to delete ``pin well''.
If an agent takes an action $A_1=1$, the updated sentence is ``Good morning, George. I hope you \textit{pin}'' and the reward is $R_1$=-1.
The updated intent is then 1.
Next, the agent take an action $A_2=3$.
The updated sentence is ``Good morning, George. I'' and the reward is $R_2=-2$.
In a real-life application we built\footnote{This application is not part of the paper.},
we observe the action that is done by the user after the agent's action.
If the agent takes $A_2=3$ the user has to re-dictate the words ``hope you''
indicating that they were deleted incorrectly.
In our simulation (described further), when we inject noisy words into a sentence, we know the amount of noise and thus
can simply calculate the reward for any given action.

\section{Model-Based Actor-Critic Algorithm}
\label{mbac-section}
\begin{figure}[t]
\centering
\includegraphics[width=0.45\textwidth]{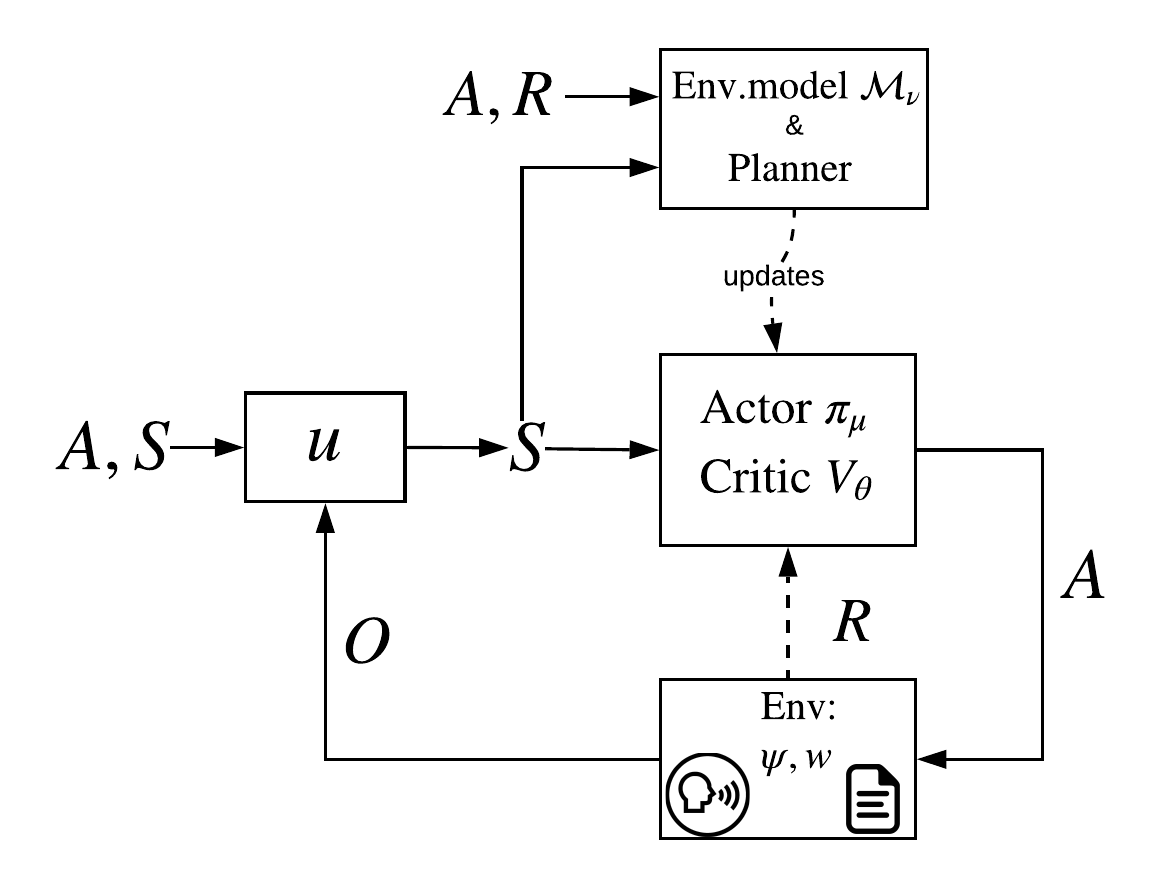}
\caption{ MBAC architecture with the primary model-based components.
Built on the work of ~\citealp{sutton2018reinforcement} (subscripts are omitted to avoid overlap of time-steps).}
\label{model-based-rl}
\end{figure}

The application of a common actor-critic model to our specific task
was motivated by its strong performance across many domains.
Actor-critic methods are one of the most stable on-policy model-free
algorithms.
Our Model-Based Actor-Critic (MBAC) algorithm (Algorithm~\ref{alg:algi}) includes major components of model-based RL architectures
following~\citeauthor{sutton2018reinforcement}.
Figure~\ref{model-based-rl} depicts these components and interaction between them: the environment
model $\mathcal{M}_{\nu}$, the state-update
function $u$, the policy or `actor' $\pi_{\mu}$, the state-value
function or `critic' $V_{\theta}$, and a planner with policy $\pi^{\text{planner}}$.
Taking the parameterized function approximation setting, we denote the parameters of the model, policy, and state-value function as $\bm\nu, \bm\mu, \bm\theta$ respectively.

Environment models are fundamental in making long-term predictions
and evaluating the consequences of actions.
People quickly learn and adjust to a course of dialogues because they
are good at those predictions.
The goal of the model  $\mathcal{M}_{\nu}$ is to learn the transition dynamics
of the environment---to predict the next states and rewards from
the previous state and last action.
Using the model, action choices are evaluated during planning by anticipating possible futures given different actions.
An~agent that uses planning and an environment model can be better in
computing long-term consequences of actions,
such as predictions about the estimated returns from states or~state-action pairs.
In other words, environment models can improve the agent's prediction and control
abilities.
Thus, model-based approaches are foundational for applications of
RL to real-life problems, such as interactive dialogues.

Instead of making the assumption that states are Markov,
i.e., states are fully-observable, we generalize to POMDP.
Together with the previously observed action, $A_{t-1}$, and previous state, $S_{t - 1}$, the observation is used by the
\textit{state-update function} to produce a state $S_t = u (S_{t-1}, A_{t-1}, O_t)$.
Incorporating a state-update function in the algorithm
is necessary in order to encapsulate the course of a dialogue into a compact
summary that is useful for choosing future actions.
For example, if a user's intent is to delete five words,
but at the first step, the agent
deletes only one word, the agent should have the information of this first attempt
before trying to delete the remaining four words.
The idea is that the state should be sufficient for predicting and controlling
the future trajectories and understanding the consequences of the actions,
{\it without having to store a complete history}.
The state $S_t$ is used as an input for the model and planner.
The model $\mathcal{M}_{\nu}$ uses this state $S_t$ together with the action taken $A_t$ and subsequently observed reward $R_{t + 1}$ and state $S_{t + 1}$ when training.
Since the model $\mathcal{M}_{\nu}$ becomes better in its predictions and the planner learns better policy, both are used to update model-free components, namely the
policy $\pi_{\mu}$ and the value function $V_{\theta}$.

Provided the necessity of the state-update function, implementations like DQN~\cite{mnih2013playing}
are not practical for this approach because it would require the storage of
a potentially long sequence of state updates in a memory replay buffer.
These implementations would be resource inefficient and contain problems of staleness as the state update function is being trained.
Thus, online on-policy policy-based gradient methods, such as actor-critic,
and in particular, advantage actor-critic~\cite{mnih2016asynchronous}
are more suitable.

For each time step $t$ of an interaction, we compute the state $S_t$ from
the previous state $S_{t-1}$, incorporating
the information that followed in the transition: the action $A_{t - 1}$ and the observation $O_{t}$ received by the agent after taking action $A_{t-1}$.
Then, we perform three major steps (see Algorithm~\ref{alg:algi}):
1) Planning and acting; 2) Updating the policy $\pi_{\mu}$ and the state-value function $V_{\theta}$;
and 3) Updating the model $\mathcal{M}_{\nu}$.
In comparison, the model-free actor-critic algorithm does not include the planning portion of step (1) (lines 6-11) and step (3).
In addition, in Step (2), we modify the updates for both the actor and critic.
\subsection{Step 1. Planning and Acting (lines 6-13)}
First, we compute the predicted next states $\hat S_{t+1}^a$ and the predicted
rewards $\hat R_{t+1}^a$ for each possible action $ a \in \mathcal{A}$,
using the model $\mathcal{M}_{\nu}$.

\begin{algorithm}[htb]
\caption{Model-Based Actor-Critic (MBAC)}\label{alg:algi}
\begin{algorithmic}[1]
\State Initialize $\mathcal{M}_{\nu}, \pi_{\mu}, V_{\theta},
                \pi^{\text{planner}}, \bm \nu, \bm \mu, \bm \theta$
\State Hyperparameters: $\alpha_{\nu, \mu, \theta}, \gamma, \beta$
\For {each episode}
    \For {t = 1...T}
    \State $S_t \gets u (S_{t - 1}, A_{t - 1}, O_{t})$
        \State \text  {\# Plan and act}
        \For {$ a \in \mathcal{A}$}
            \State $\hat S_{t+1}^a, \hat R_{t+1}^a \gets \mathcal{M}_{\nu}(S_t, a)$
            \State $\hat G_t^a = \hat R_{t+1}^a + \gamma V_{\theta} (\hat S_{t+1}^a)$
        \EndFor
        \State $\pi ^{\text {planner}} \gets \text{SoftMax}(\widehat {\mathbf{G_t}})$
        \State $V^{\text {planner}}(S_t) \gets \sum\limits_{a \in \mathcal{A}} \hat G_t^a \pi^{\text {planner}}(a | S_t)$
        \State Sample action $ A_t \sim \pi^{\text {planner}}(\cdot | S_t)$
        \State Take action $A_t$ and get $O_{t + 1}$ and $R_{t + 1}$
        \State \text{\# Update actor $\pi_{\mu}$ and critic $V_{\theta}$}
         \State $\mathcal{L}^{\text {critic}} \gets (V_{\theta}(S_t) - V^{\text {planner}}(S_t) )^2$\\
        \State $\mathcal{L}^{\text{actor}} \gets  \sum\limits_{a \in \mathcal{A}}                     \pi_{\mu}(a | S_t) \log
                    \frac{\pi_{\mu}(a | S_t)}{\pi^{\text {planner}}(a | S_t)}$\\
        \State \small $H \gets
                 -\sum\limits_{a \in \mathcal{A}} \pi^{\mu}(a | S_t)
                \log \pi^{ \mu}(a | S_t)$\\
        \State \normalsize  $\mathcal{L}^{\text{actor}} \gets  \mathcal{L}^{\text{actor}} - \beta H$
        \State $\bm\theta \gets \bm\theta + \alpha_{\theta} \nabla_{\theta}\mathcal{L}^{\text {critic}}$
        \State $\bm\mu \gets \bm\mu + \alpha_{\mu} \nabla_{\mu}\mathcal{L}^{\text {actor}}$
        \State \text{\# Update model $\mathcal{M}_\nu$}
        \State $ S_{t+1} \gets u (S_t, A_t, O_{t+1})$
        \State $\hat S_{t+1}, \hat R_{t+1} \gets \mathcal{M}_{\nu}(S_t, A_t)$
        \State $\mathcal{L}^{\text {state}} \gets  ||\hat S_{t+1} - S_{t+1}||_2$
        \State $\mathcal{L}^{\text {reward}} \gets  (\hat R_{t+1} - R_{t+1})^2$

        \State $\mathcal{L} \gets \mathcal{L}^{\text {reward}} + \mathcal{L}^{\text {state}}$
        \State $\bm\nu \gets \bm \nu + \alpha_{\nu} \nabla_{\nu} \mathcal{L} $
        \EndFor
\EndFor
\end{algorithmic}
\end{algorithm}

Using the predicted next states $\hat S_{t+1}^a$ and rewards $\hat R_{t+1}^a$,
we then compute a set of estimated discounted returns.
While we can iterate the model for multiple steps at the cost of more computation, we found that one-step iteration was adequate.
Thus, we use the value estimate of the one-step predicted state:
$
    \hat G_t^a = \hat R_{t+1}^a + \gamma V_\theta(\hat S_{t + 1}^a)$
where $\gamma$ is a hyper-parameter, $ 0 \leq \gamma \leq 1$ called
the~\textit{discount rate}.

We compute the planner policy $\pi ^{\text {planner}}$
by applying the softmax function to the vector of estimated returns $\widehat {\mathbf{G_t}}$.
The state-value corresponding to the policy of the planner is computed by:
\begin{equation}
    V^{\text {planner}}(S_t) = \sum\limits_{a \in \mathcal{A}} \hat G_t^a \cdot \pi^{\text {planner}}(a | S_t).
\end{equation}
Using the planner policy $\pi ^{\text {planner}}$, the agent takes an action $A_t$ and receives
reward $R_{t+1}$, followed by the observation $O_{t+1}$.
\subsection{Step 2. Updating Actor and Critic (lines 14-23)}
We compute the objective $\mathcal{L}^{\text {critic}}$ for the critic
using the mean square error (MSE) between the
state-value of the critic $V_{\theta}(S_t)$ and the state-value of the
planner $V^{\text {planner}}(S_t)$, which was based on the model reward predictions and critic's estimates at predicted states:
\begin{equation}
    \mathcal{L}^{\text {critic}} = ( (V_{\theta}(S_t) - V^{\text {planner}}(S_t))^2.
\end{equation}
The critic is updated at each time step by the following rule:
\begin{equation}
    \bm\theta_{t+1} = \bm\theta_{t} + \alpha_{\theta} \nabla_{\theta}\mathcal{L}^{\text {critic}}_{t}.
\end{equation}
The actor $\mathcal{L}^{\text{actor}}$ objective is computed by using
the relative entropy between the actor policy $\pi_{\mu}$ and the
planner policy $\pi^{\text {planner}}$:
\begin{equation}
\label{six}
    \resizebox{0.42\textwidth}{!}{%
    $
    \mathcal{L}^{\text{actor}} (t)=  \sum\limits_{a \in \mathcal{A}} \pi_{\mu}(a | S_t) \log \frac{\pi_{\mu}(a | S_t)}{\pi^{\text {planner}}(a | S_t)} - \beta H $
    }
\end{equation}
where $H$ is an entropy term being
subtracted to encourage exploration similarly to~\citealp{mnih2016asynchronous}.
The entropy term $H$ in Equation~\ref{six} is
defined as:
\begin{equation}
    \resizebox{0.42\textwidth}{!}{
    $
    H (\pi_{\mu}(a | S_t)) =
        -\sum\limits_{a \in \mathcal{A}} \pi_{\mu}(a | S_t)
        \log \pi_{\mu}(a | S_t) $.
        }
\end{equation}
The actor is updated similarly to the critic at each time step by the following rule:
\begin{equation}
    \bm\mu_{t+1} = \bm\mu_{t} + \alpha_{\mu} \nabla_{\mu}\mathcal{L}^{\text {actor}}_{t}.
\end{equation}

\begin{figure}[t]
\centering
\includegraphics[width=0.5\textwidth]{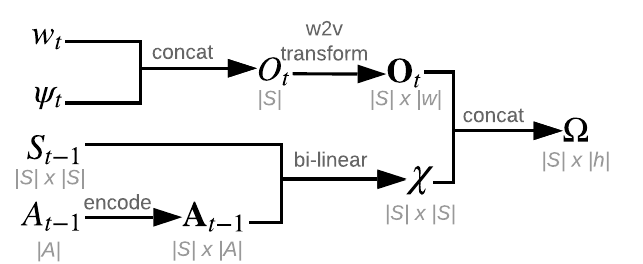}
\caption{ {Construction of RNN-input $\Omega$ for the state-update function. The $\Omega$ includes the information of the previous state, the observation, and the reward.
The dimensions of each array are shown underneath each entity.}}
\label{rnn-input}
\end{figure}
\subsection{Step 3. Model Update (lines 24-30)}
We update the model by minimizing the state-objective $\mathcal{L}^{\text {state}}$
and the reward-objective $\mathcal{L}^{\text {reward}}$ together.
Using the current state $S_t$, the action taken $A_t$, and the new observation
$O_{t+1}$ as an input for the state-update function $u$, we compute the next state
$S_{t+1}$. We also compute the predicted next state $\hat S_{t+1}$ and reward $\hat R_{t+1}$
by using the model $\mathcal{M}_{\nu}$, given the current state $S_t$ and action taken $A_t$.
The state-objective $\mathcal{L}^{\text {state}}$ is how correct our model
is in the state-prediction:
\begin{equation}
    \mathcal{L}^{\text {state}} = \|\hat S_{t+1} - S_{t+1}\|_2.
\end{equation}
Similarly, using the MSE between the predicted and
actual rewards, we get the reward-objective $\mathcal{L}^{\text {reward}}$.
Adding these two objectives together, we can then update the model parameters:
\begin{equation}
    \bm\nu_{t+1} = \bm\nu_{t} + \alpha_{\nu} \nabla_{\nu}(\mathcal{L}^{\text {state}} + \mathcal{L}^{\text {reward}})_{t}.
\end{equation}

\section{Implementation}
\label{implem}
\subsection{Architecture}
The state-update function $u$ is represented by a recurrent neural network (RNN): a two-layered bi-directional GRU~\cite{cho2014learning} with input
and hidden sizes of 400 and 100, respectively.
In order to compute an updated state using the RNN, we need to provide an~\textit{RNN-input} $\Omega$, which includes the previous state, the observation, and the reward (Figure~\ref{rnn-input}).
The observation $O_t$ is computed by concatenation of the encoded text $w_t$ and user input $\psi_t$.
\begin{figure}[t]
\centering
\includegraphics[width=0.5\textwidth]{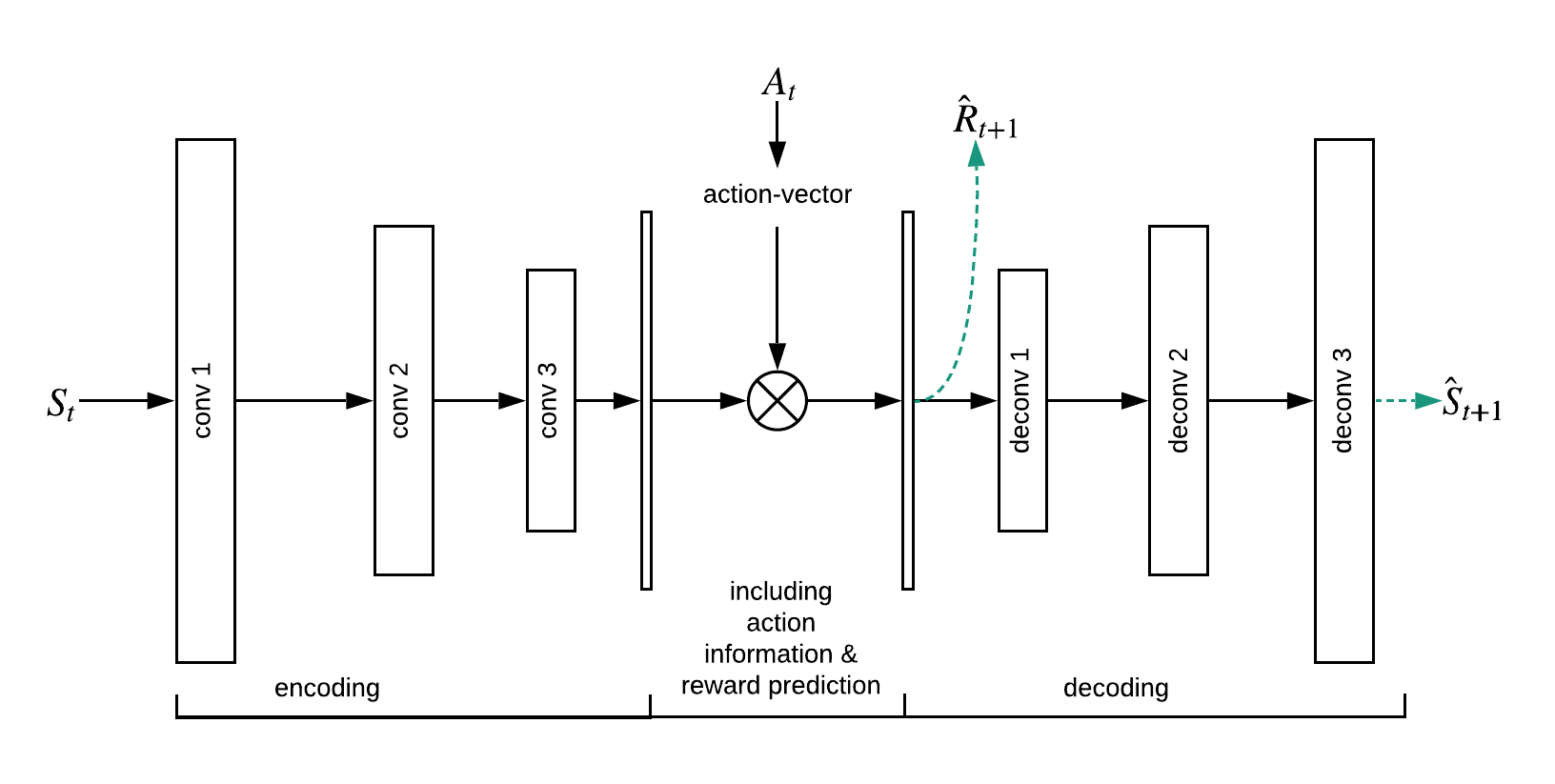}  
\caption{ {A model $\mathcal{M}$ architecture.
The state $S_t$ and the action $A_t$
are inputs to the model and
the model computes predictions of the next reward $\hat R_{t+1}$
and the next state $\hat S_{t+1}$.}}
\label{model}
\end{figure}
We use word2vec
word embeddings~\cite{mikolov2013distributed} to encode
$O_t$ into an~\textit{observation matrix} $ \bm O_t$.
An action is encoded into an~\textit{action-matrix} $\bm A_{t-1}$.
A bi-linear transformation of the form
$ S_{t-1}Z\bm{A_{t-1}}+b$ is applied to the
action-matrix and the previous state to
compute~\textit{previous-state-action information} $\chi$.
The previous-state-action information $\chi$ is concatenated
with the observation $\bm O_t$ to compute $\Omega$.
We note that using one of the latest developments such as BERT~\cite{devlin2018bert} for word embeddings could not apply to our settings.
BERT models jointly condition on both the left and right contexts of a sentence,
which would create a problem in our simulation provided we inject random noise.
Instead, we learn temporal structures with an RNN.

We use single-threaded agents that learn with a single stream of experience.
The use of single-threaded agent mimics real human-computer interaction for the task.
The actor and critic are a 1D convolutional neural network
that extracts correlations in the temporal structure of the sentences.
The network is shared
up until the point where there is a split into two separate heads:
one producing a scalar value estimate and the other a probability distribution over actions as a policy.
The shared architecture consists of three convolutional layers with filter sizes of 50, 50, and 100.
The entropy term $H$ was used in both the advantage actor-critic and the MBAC architectures.

The model architecture (Figure~\ref{model}) for state prediction
is similar to~\cite{oh2015action}.
We use a stack of three convolutional layers with filter sizes 50, 50, and 100.
The action is encoded into a vector and Hadamard product is applied to the output of the 1D
convolutional layers with the action vector.
The output of this operation is then passed through a stack of three
deconvolutional layers with filters sizes of 100, 50, and 100. The output of the Hadamard product operation is also passed through a linear layer to a single node for the scalar reward prediction.
The state $S_t$ and the action $A_t$
are inputs to the model and
the model outputs predictions of the next reward $\hat R_{t+1}$
and the next state $\hat S_{t+1}$.
\begin{figure}[]
\centering
\includegraphics[width=0.5\textwidth]{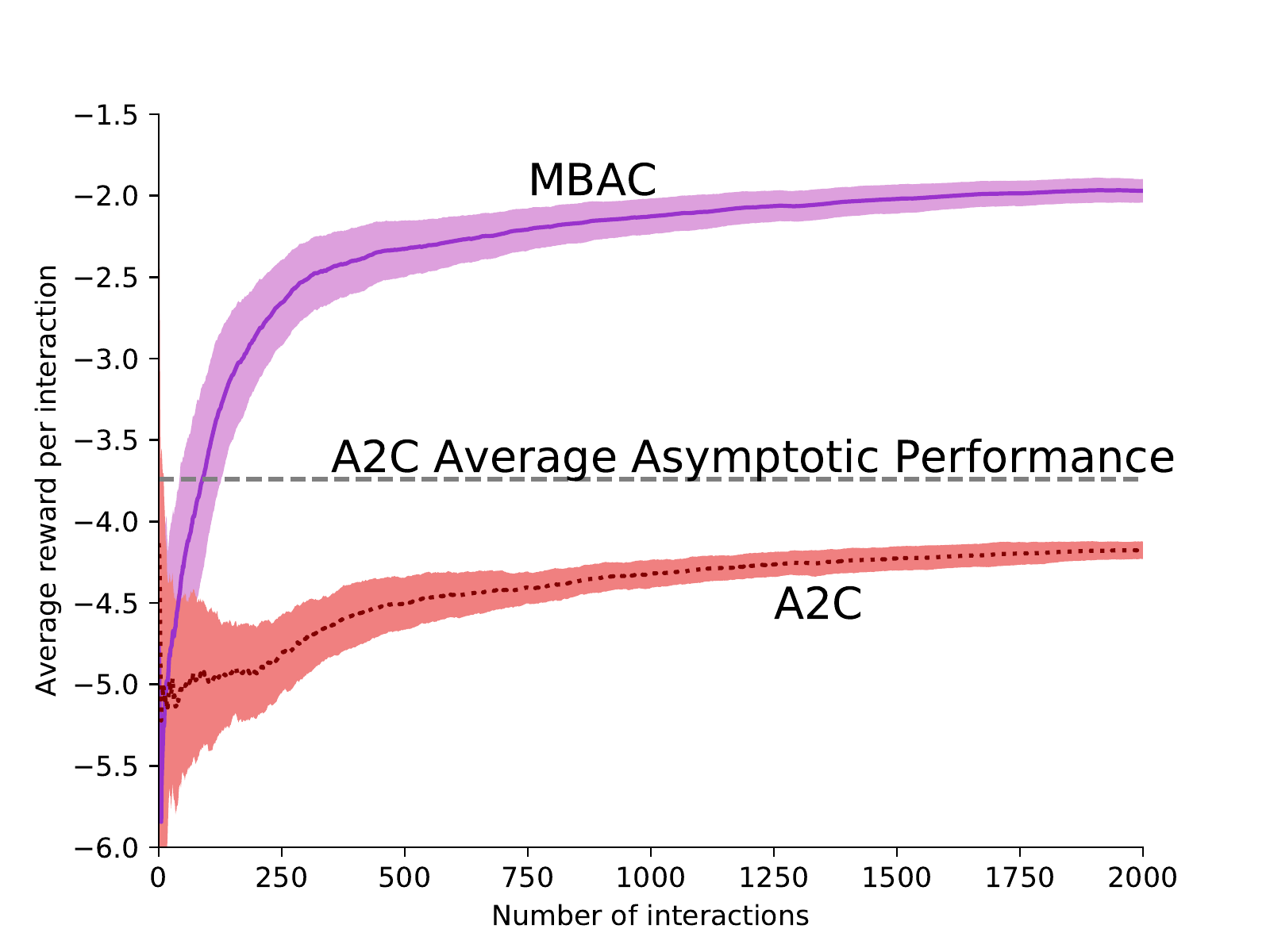} 
\caption{ {Average reward per interaction for MBAC and A2C: it takes MBAC 70 times less samples to achieve the average reward at which A2C approximately converges.}}
\label{avg-reward}
\end{figure}
\subsection{Initialization \& Hyperparameters}
The architecture was implemented in PyTorch 1.0.
The RNN-input requires an initial state and an initial action for a state construction.
The intial state was initialized to ones, and the initial action
was initialized to zeros.
The actions were represented by the range $[1, 15]$, respecting the real-world
setting, where a user is most likely to observe the incorrectly recognized words
and therefore halt dictation, before a sentence becomes longer than 15 words.
We used the Adam optimizer~\cite{kingma2014adam} with the default parameters, with
the exception of the learning rate, which was set to $1\text{x}10^{-4}$.
We used a discount rate $\gamma$ of 0.9, and clipped gradients at 0.9.
The word-embedding size was 300.
Entropy parameter $\beta$ was set to 1.
\subsection{Simulation}
\label{sim}
The simulation of the chosen document-editing dialogue task required
mimicking a sentence that was corrupted by the voice-recognition component.
We chose the BookCorpus dataset~\cite{zhu2015aligning}.
Along with narratives, the books contain dialogue, emotion, and a wide
range of interaction between characters~\cite{kiros2015skip}.
The sentences from the dataset are fed to the model one by one,
and noisy words\footnote{\small {To simulate noise phonetically close to words would require significant engineering efforts (such as creating datasets of phonemic dictionaries) that is beyond the scope of the present work.}}
are injected at a random point in a sentence.
This intentionally creates a difficult problem for the agent:
there are no labels, and the noise is random.
Thus the only way for the agent to learn is through dialogue
interaction and by observing the dynamics of the environment.

\begin{figure}[]
\centering
\includegraphics[width=0.45\textwidth]{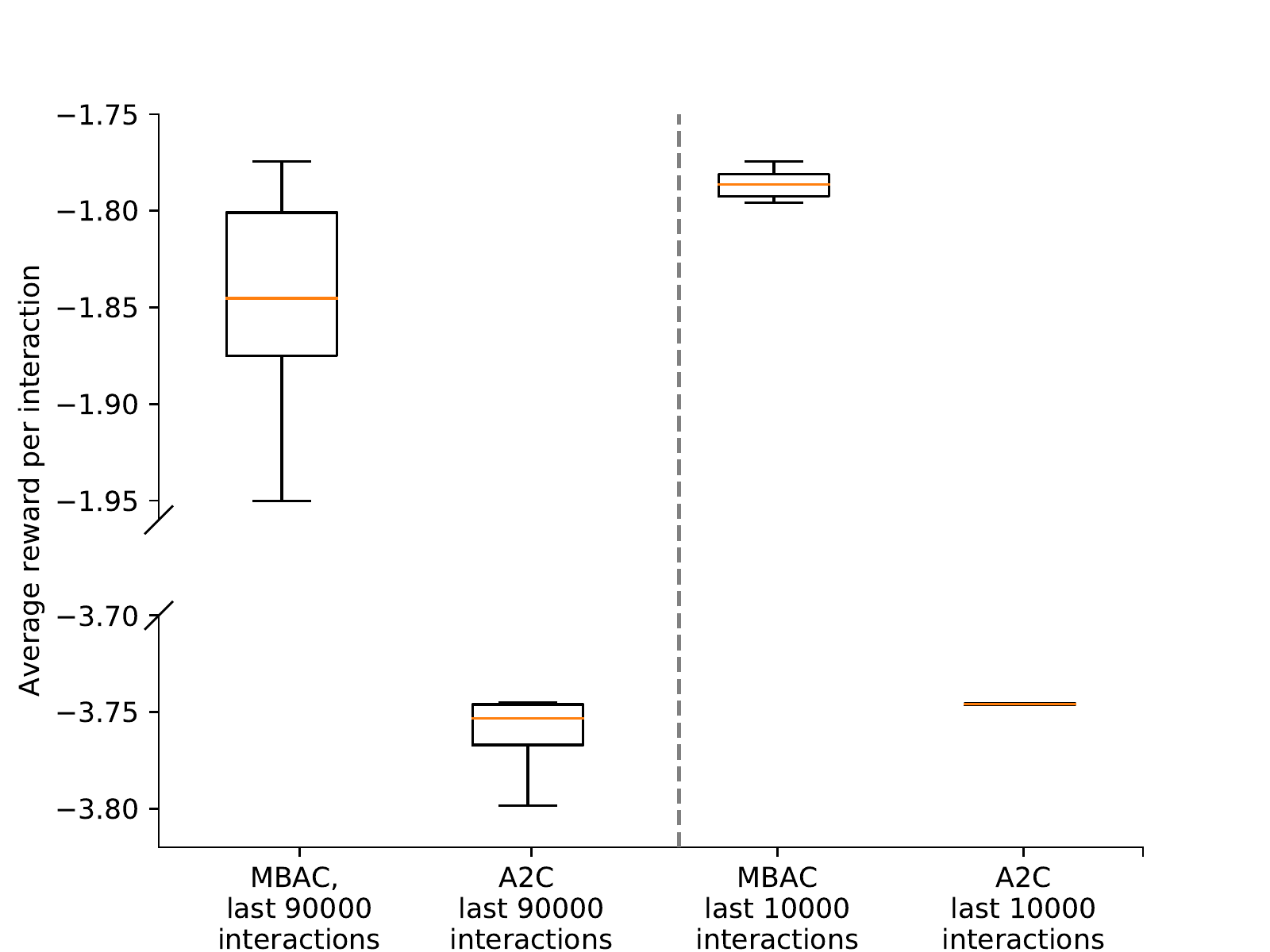}
\caption{ {Asymptotic performance over the last 90,000 and 10,000 interactions for MBAC and A2C: MBAC is just over 2 times better in the limit. Orange line is the median.}}
\label{avg-reward-box}
\end{figure}
\section{Results}
The simulation assumes that the agent always receives
``No'' as a speech input $\psi$
from the user, provided we are interested in the interaction itself---when the agent has to make an action, and not in idle moments when the
dictation is going well and all words are well recognized.

In the experiments, each algorithm was executed for 100,000 simulated interactions, with 30 runs for each algorithm, each time changing the random seed.
The random seed affects the initialization of parameters of the neural networks,
and the noise injected in the sentences.
The average over that 30 runs was used in the
performance measures of the algorithms.

We demonstrate a few measures of the performance: average reward per each interaction, absolute error per step, and the distribution over actions.
\subsection{Sample Efficiency}
We first ask:
how much can model-based RL improve the efficiency
of model-free deep RL algorithms?
In other words, does the MBAC algorithm
perform better than the ordinary actor-critic methods,
without the model and the planner?

Figure~\ref{avg-reward} shows that MBAC takes about 70
steps on average to achieve an average reward of -4.
Even at 2,000 A2C does not reach its asymptotic performance while MBAC shows twice higher average reward per interaction, using the same learning rate.
\begin{figure}[]
\centering
\includegraphics[width=0.5\textwidth]{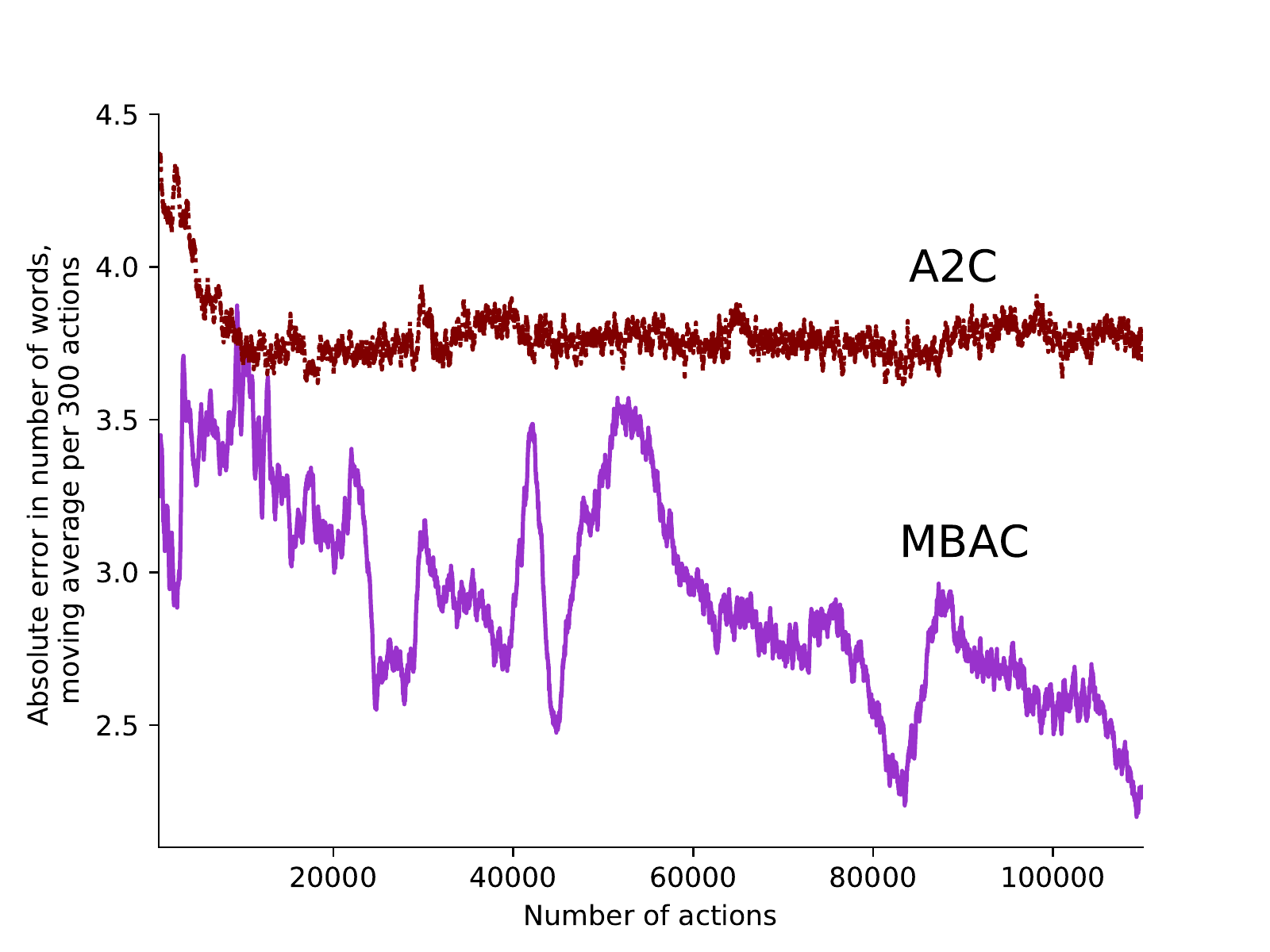}
\caption{ {Absolute error in number of words deleted for each action, averaged over 300 actions. MBAC has the lowest error. Note, that one interaction can include a few actions.}}
\label{error}
\end{figure}
It takes A2C over 71,900 steps to reach its asymptotic performance of -3.74,
while it takes MBAC only 88 steps to reach this point.
This makes MBAC over 70 times more sample-efficient.
Moreover, MBAC quickly gains a higher average reward and continues to improve
while A2C plateaus.

Provided the implementation of single-threaded agents for MBAC algorithm,
we do not compare to Asynchronous Advantage Actor-Critic, or A3C.
A3C is the asynchronous version of A2C which uses data samples
from several asynchronous environments simultaneously and we believe
that comparison of single-threaded agents is foundational as a starting point.

\subsection{Asymptotic Performance}

MBAC also achieves just over 2 times the performance of standard A2C asymptotically.
Figure~\ref{avg-reward-box} shows performance over the last 90,000 and last 10,000
interactions.
MBAC performance improves from -1.84 to -1.77 while A2C performance only improves from -3.76 to -3.74, which remains almost the same.

\begin{figure}[]
\centering
\includegraphics[width=0.5\textwidth]{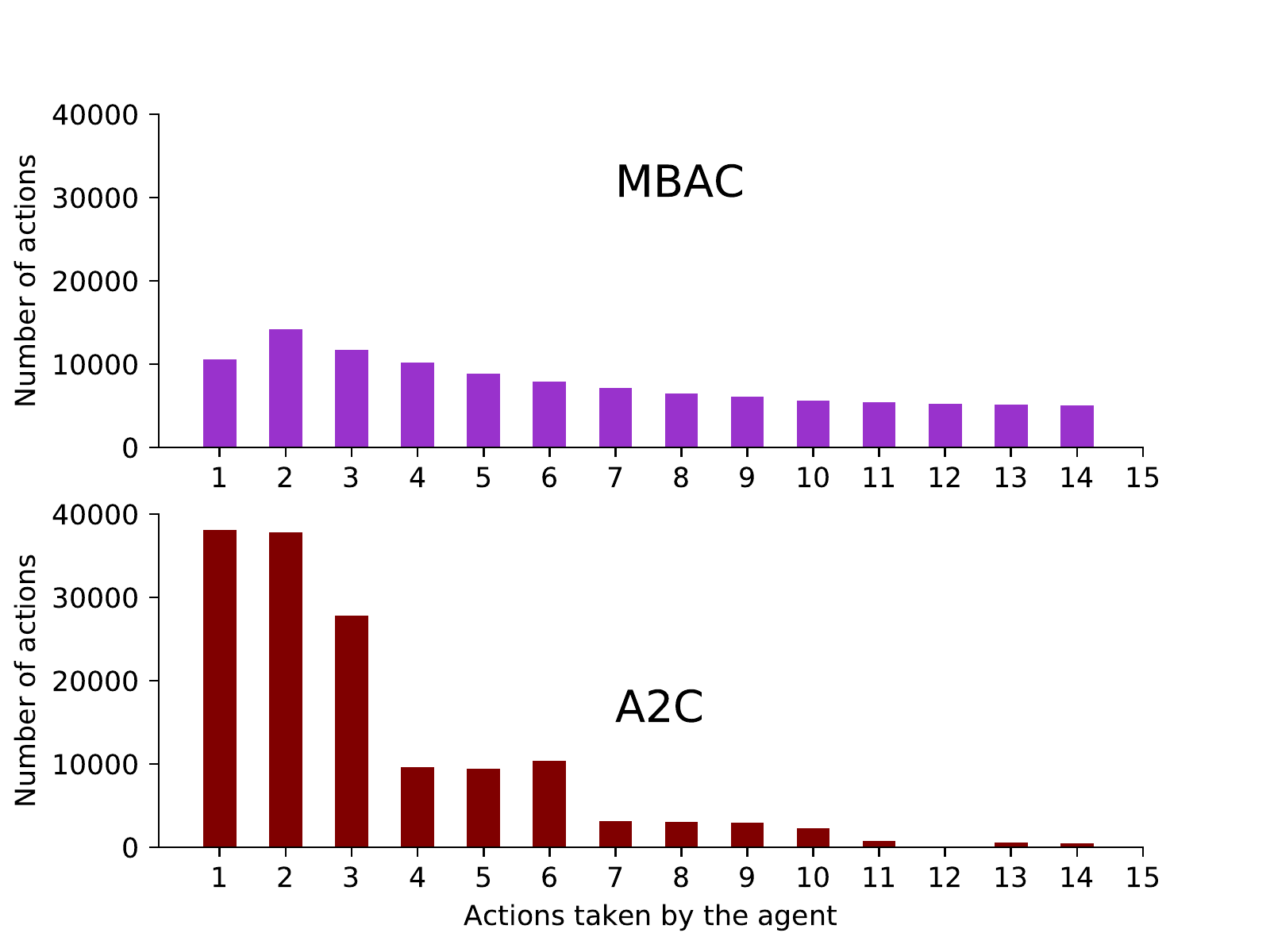}
\caption{ {The distribution over actions shows that MBAC is less deterministic than A2C.}}
\label{actions-train}
\end{figure}

We believe MBAC's performance is explained by improved exploration.
Figure~\ref{actions-train} demonstrates the distribution of action selection by the agent
for both MBAC and A2C.
A2C prefers lower numbered actions, and explores less, while MBAC's distribution
over actions is much more uniform.

Figure~\ref{error} shows the absolute error in the number of words---how far the agent was
from the intent on every action.
We found this is important to observe in addition to the average reward, as the latter only provides us
with the reward information per interaction.
Similar to the average reward, MBAC's absolute error keeps decreasing, while A2C's absolute error
remains within the interval $[3.5,4.2]$.

\begin{figure}[h]
\centering
\includegraphics[width=0.5\textwidth]{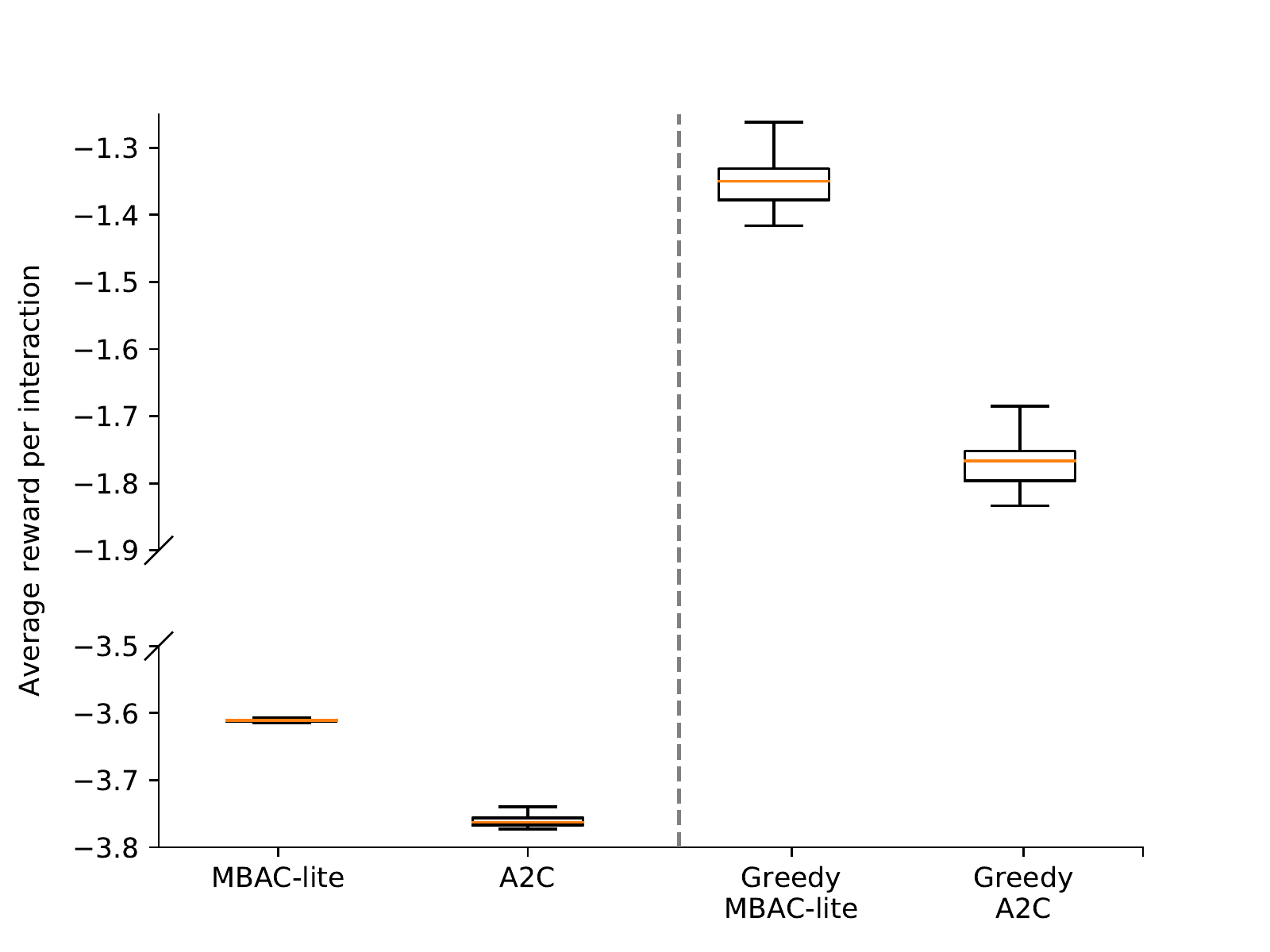}
\caption{ {The performance of MBAC-lite vs.~A2C. MBAC-lite outperforms A2C with and without applying greedy actions. The orange line is the median and the extent of the boxes represent upper and lower quartiles.}}
\label{eval}
\end{figure}

Finally, motivated by the future deployment of this solution into a real-life
application, we demonstrate that using the model-free portion of the MBAC architecture
would still work and allow for deployment and implementation that is less computationally and memory heavy.
In other words, can we first apply MBAC training, then detach the model and planner, so as to deploy only the model-free architecture?
We do exactly this and call the resulting model-free agent {\it MBAC-lite}. For evaluation,
we use a subset of the dataset of over 74 million sentences
that was split into halves for training and testing.
We used the test set of unseen sentences to assess performance.
Figure~\ref{eval} compares the performance of two model-free architectures: MBAC-lite and A2C.
Two results emerged: 1) MBAC-lite can be easily
detached in order to be deployed more efficiently with performance superior to A2C; and
2) Using greedy actions\footnote{Selecting an action whose estimated value is the greatest, instead of sampling from a probability distribution, which is done in non-greedy settings.}, MBAC-lite also performs significantly better than A2C.
This experiment demonstrates how we can distill model and planner knowledge into the model-free architecture.


\section{Conclusions}
In this work, we presented a sample-efficient
model-based actor-critic algorithm.
At deployment time, the model and planner can be detached to obtain an inexpensive
yet high performing model-free agent.
Our experiments demonstrated that MBAC learned 70 times faster and obtained 2 times the performance
of classic A2C in just under 100 interactions.
We believe this result is foundational in applying
reinforcement learning to human-computer interactions.

In this work, we presented a novel update to a model-free policy using a soft-planner policy.
We see studying various methods of updating agents' model-free components with models and planners as a fruitful research direction.
As a long-term vision, we believe that model-based RL
is a promising and more efficient alternative to model-free RL.

\section{Acknowledgments}
We would like to thank the following collaborators for their insightful conversations:
J. Fernando Hernandez-Garcia, Hado van Hasselt, Kris De Asis, Matthew Schlegel,
Niko Yasui, Patrick Pilarski,  and Richard S. Sutton.

\end{document}